\documentclass[journal]{IEEEtran}
\IEEEoverridecommandlockouts
\usepackage{amsmath,amssymb,amsfonts}
\usepackage{algorithm}
\usepackage{algorithmic}
\usepackage{graphicx}
\usepackage{textcomp}
\usepackage{csquotes}
\usepackage{xcolor,soul}
\usepackage{adjustbox}
\usepackage{hyperref}
\usepackage{multirow}
\usepackage{subcaption}
\usepackage{makecell}
\usepackage{float}
\usepackage{authblk}
\usepackage{orcidlink}

\usepackage[numbers,sort&compress]{natbib}

\usepackage{etoolbox}

\usepackage{amsmath}

\usepackage{mathtools}
\usepackage{gensymb}
\usepackage{enumitem}
\usepackage{diagbox}
\usepackage{parskip}

\makeatletter
\newcommand{\thickhline}{%
    \noalign {\ifnum 0=`}\fi \hrule height 1.5pt
    \futurelet \reserved@a \@xhline
}

\begin{document}
\title{Neural-Assisted in-Motion Self-Heading Alignment}


\author{\IEEEauthorblockN{Zeev Yampolsky$^1$\IEEEauthorrefmark{1}\orcidlink{0009-0003-9122-7576}, Felipe O. Silva$^2$ \orcidlink{0000-0002-3715-7023}, Adriano Frutuoso$^3$\orcidlink{0000-0002-3907-9815}, and Itzik Klein$^1$\orcidlink{0000-0001-7846-0654}}\break  
\and

\IEEEauthorblockA{$^1$The Hatter Department of Marine Technologies,\break 
{Charney School of Marine Sciences, University of Haifa, Haifa, Israel}}
\and 

\IEEEauthorblockA{$^2$Department of {Automatics, School of Engineering} \break
{Federal University of Lavras, Lavras, MG, 37200-900, Brazil}}
\and
\IEEEauthorblockA{ \break{$^3$Department of {Higher Education} \break
{Federal Institute of Amazonas, Manaus, AM, 69075-351, Brazil}}}
\thanks{\IEEEauthorrefmark{1}Corresponding author: Zeev Yampolsky, zyampols@campus.haifa.ac.il.}}

\maketitle
\begin{abstract}
    Autonomous platforms operating in the oceans require accurate navigation to successfully complete their mission. In this regard, the initial heading estimation accuracy and the time required to achieve it play a critical role. The initial heading is traditionally estimated by model-based approaches employing orientation decomposition. However, methods such as the dual vector decomposition and optimized attitude decomposition achieve satisfactory heading accuracy only after long alignment times. To allow rapid and accurate initial heading estimation, we propose an end-to-end, model-free, neural-assisted framework using the same inputs as the model-based approaches. Our proposed approach was trained and evaluated on real-world dataset captured by an autonomous surface vehicle. Our approach shows a significant accuracy improvement over the model-based approaches achieving an average absolute error improvement of 53\%. Additionally, our proposed approach was able to reduce the alignment time by up to 67\%. Thus, by employing our proposed approach, the reduction in alignment time and improved accuracy allow for a shorter deployment time of an autonomous platform and increased navigation accuracy during the mission.
    \end{abstract}

\section{Introduction}\label{intro_sec}
Accurate initial attitude alignment is a fundamental prerequisite for the proper operation of strapdown inertial navigation systems (SINS), directly affecting the performance of subsequent navigation, guidance, and control tasks \cite{Silva_2016,farrell_2022}. Among the attitude components, heading (yaw) alignment is widely recognized as the most challenging, particularly in scenarios where the vehicle experiences limited excitation, low dynamics, or operates in environments with poor (or none) external aiding \cite{groves}. This challenge is especially pronounced in marine and underwater applications, such as autonomous underwater vehicles (AUVs), where prolonged stationary or quasi-stationary conditions, slow drift, and low signal-to-noise ratios are common \cite{Gu_2008,Sun_2010}.\\
Classical approaches to coarse self-alignment rely on analytical or optimization-based formulations that exploit vector observations derived from inertial sensor measurements and known navigation-frame quantities, such as gravity and Earth rotation \cite{Britting_1971}. Seminal works on attitude determination and alignment, including Wahba’s problem formulation \cite{Wahba_1966}, early analytical alignment methods \cite{Thompson_1966, Krishnan_1970, Jiang_1998}, and subsequent developments in observability analysis \cite{Hermann_1977,Wu_2012,Rothman_2014}, have established a solid theoretical foundation for SINS alignment. In practice, modern coarse alignment techniques, including variants of the attitude decomposition-based initial alignment (ADIA) and the optimization-based ADIA (OBA), construct observation vectors through temporal integration or averaging of inertial measurements, followed by closed-form or iterative attitude estimation \cite{Silson_2011,Wu_2011,wu_2013}.\\
Despite their strong theoretical grounding, the performance of these classical methods is tightly coupled to the quality and richness of the excitation experienced during the alignment interval. Recent experimental studies have highlighted the practical limitations of such approaches under realistic operational conditions. In \cite{frutuoso_2023_oceaneng}, a comprehensive performance evaluation of coarse alignment methods for AUVs operating under mooring conditions demonstrated that even carefully tuned analytical methods may suffer from significant accuracy degradation due to environmental disturbances, sensor noise, and limited observability. Extending this analysis, the results reported in \cite{frutuoso_2024} showed that under real drift conditions, where excitation is weak and highly non-stationary, the variability in alignment performance increases substantially, and the benefits of more sophisticated analytical formulations become marginal. More recently, the controlled in-motion alignment experiments presented in \cite{frutuoso_2025} further confirmed that, while deliberate maneuvers can improve observability, the relationship between motion patterns and achievable heading accuracy remains complex and difficult to exploit optimally using fixed analytical models.\\
These findings suggest that, although classical self-alignment techniques are physically interpretable, their effectiveness in practice is constrained by assumptions on excitation, noise characteristics, and model fidelity that are often violated in real-world operations. This has motivated growing interest in data-driven approaches for inertial navigation, particularly those based on machine learning (ML) and neural networks (NN), which offer the potential to learn complex nonlinear relationships directly from sensor data.\\
In recent years, the introduction of data-driven frameworks, such as neural networks (NN), has been increasing in the navigation literature \cite{cohen_2024,chen2024deep}. Neural networks in the navigation literature are usually divided into end-to-end approaches, where the NN replaces the entire model and processing pipeline. These models are sometimes referred to as model-free frameworks. An additional category is hybrid models, where the NN is incorporated into the navigation pipeline. Research shows that relatively simple NN architectures, such as one-dimensional and two-dimensional convolutional neural networks (1DCNN and 2DCNN), or different types of recurrent neural networks (RNN), demonstrate impressive results in tasks such as calibration, gyrocompassing, and more \cite{yampolsky2025dcnet,huang2022mems,stolero2026rapid,versano2026wminet,vertzberger2021attitude,mu2019end,zhang2025data}. More advanced NN architectures have also been utilized to estimate the process noise of an extended Kalman filter \cite{cohen2025adaptive} by employing a transformer architecture \cite{vaswani2017attention}.\\
Motivated by these observations, this paper proposes an end-to-end neural-assisted in-motion self-heading alignment method that combines the physical structure of classical inertial alignment with the inference power of deep neural networks. Instead of explicitly constructing analytical observation vectors or solving an attitude determination problem, the proposed neural network approach employs a 2D convolutional layer-based framework to estimate the heading angle of a vessel. The neural network architecture is deliberately designed to reflect the physical decomposition of the alignment problem by employing a multi-head structure. The proposed architecture leverages the noise reduction and regression capabilities of neural networks \cite{cong2023review} to estimate the heading angle accurately and efficiently. The main contributions of this work are:
\begin{enumerate}
    \item \textbf{A neural-assisted model-free approach}: We offer a neural end-to-end approach for self-heading alignment capable of improving the initial heading accuracy and reducing the required alignment duration.
    \item \textbf{Novel multi-head CNN-based NN}: We propose a novel multi-head 2DCNN-based framework consisting of three 2DCNN heads followed by a fully connected (FC) regression block for accurate and efficient heading estimation using inertial measurements only.
\end{enumerate}
The proposed method does not aim to discard established alignment theory; rather, it builds upon insights gained from previous analytical and experimental studies, including the observed sensitivity of classical methods to excitation and integration time \cite{frutuoso_2023_oceaneng, frutuoso_2024, frutuoso_2025}. By learning an implicit mapping from physically meaningful sensor inputs to the heading estimate, the network is expected to capture higher-order temporal correlations and residual observability patterns that are difficult to exploit using fixed analytical formulations. To train and validate the proposed approach, a real-world dataset was recorded using an autonomous surface vehicle (ASV) under mooring conditions. Five experiments were conducted on five different days, resulting in a diverse real-world dataset. The dataset was divided into training and validation sets, with 35 minutes used for training and four trajectories with a total duration of 8 minutes used for evaluation. The results show an average improvement of $53\%$ over the model-based baseline approaches, while reducing the required alignment time by up to $67\%$.\\
The rest of this paper is organized as follows: first, Section \ref{prob_form_sec} provides the background for coarse alignment and describes the mathematical formulation for the baseline state-of-the-art approaches, the ADIA, DVA, and OBA approaches. Second, Section \ref{prop_approach} provides the neural network mathematical formulations and describes the proposed NN architecture. Lastly, Section \ref{res_sec} describes the real-world data used to train and validate the proposed approach and discusses the results of the proposed NN in-motion self-alignment framework.

\section{Problem Formulation and Classical In-Motion Alignment Methods}
\label{prob_form_sec}
This section presents the mathematical formulation of the inertial coarse self-alignment problem and reviews the classical ADIA methods employed for in-motion heading estimation. The presentation closely follows the notation, assumptions, and structure adopted in \cite{frutuoso_2023_oceaneng, frutuoso_2024, frutuoso_2025}, serving as a baseline for the data-driven approach proposed in this work.

\subsection{Reference Frames and Attitude Decomposition}
Let $\mathcal{F}_b$, $\mathcal{F}_n$, and $\mathcal{F}_i$ denote the body, navigation, and inertial reference frames, respectively. The vehicle attitude is represented by the Direction Cosine Matrix (DCM) $\mathbf{C}^n_b(t)$, which maps vectors expressed in $\mathcal{F}_b$ into $\mathcal{F}_n$.

Following classical inertial alignment theory \cite{Qin_2005,Silva_2016}, the attitude matrix can be decomposed as
\begin{equation}
\mathbf{C}^n_b(t)
=
\mathbf{C}^n_{n_0}(t)\,
\mathbf{C}^{n_0}_{b_0}\,
\mathbf{C}^{b_0}_b(t),
\label{eq:att_decomposition}
\end{equation}
where $\mathbf{C}^{n_0}_{b_0}$ is the constant initial attitude matrix to be estimated, and $\mathbf{C}^n_{n_0}(t)$ and $\mathbf{C}^{b_0}_b(t)$ describe the known attitude evolution of the navigation and body frames over the alignment interval $[t_0,t]$.

The matrices $\mathbf{C}^{b_0}_b(t)$ and $\mathbf{C}^{n_0}_n(t)$ are computed from the integration of angular rates as \cite{frutuoso_2025}
\begin{align}
\mathbf{C}^{b_0}_b(t) &=
\mathbf{I}
+ \frac{\sin\|\boldsymbol{\phi}(t)\|}{\|\boldsymbol{\phi}(t)\|}[\boldsymbol{\phi}(t)\times]
+ \frac{1-\cos\|\boldsymbol{\phi}(t)\|}{\|\boldsymbol{\phi}(t)\|^2}[\boldsymbol{\phi}(t)\times]^2,
\label{eq:Cbb0}\\
\mathbf{C}^{n_0}_n(t) &=
\mathbf{I}
+ \frac{\sin\|\boldsymbol{\theta}(t)\|}{\|\boldsymbol{\theta}(t)\|}[\boldsymbol{\theta}(t)\times]
+ \frac{1-\cos\|\boldsymbol{\theta}(t)\|}{\|\boldsymbol{\theta}(t)\|^2}[\boldsymbol{\theta}(t)\times]^2,
\label{eq:Cnn0}
\end{align}
with $\mathbf{C}^n_{n_0}(t)=\mathbf{C}^{n_0}_n(t)^\top$.

The rotation vectors $\boldsymbol{\phi}(t)$ and $\boldsymbol{\theta}(t)$ are computed as
\begin{align}
\boldsymbol{\phi}(t) &= \int_{t_0}^{t} \boldsymbol{\omega}^b_{ib}(\tau)\,d\tau
+ \frac{1}{2}\int_{t_0}^{t}\!\!\left(
\int_{t_0}^{\tau}\boldsymbol{\omega}^b_{ib}(s)\,ds
\times
\boldsymbol{\omega}^b_{ib}(\tau)
\right)d\tau,
\label{eq:Phi}\\
\boldsymbol{\theta}(t) &= \int_{t_0}^{t} \boldsymbol{\omega}^n_{in}(\tau)\,d\tau,
\label{eq:Theta}
\end{align}
where $\boldsymbol{\omega}^b_{ib}$ is the gyro-measured angular rate and $\boldsymbol{\omega}^n_{in}$ is the navigation frame angular rate with respect to the inertial frame.

\subsection{Attitude Decomposition-based Initial Alignment (ADIA)}

Under low-dynamic conditions, typical of mooring, drift, or controlled in-motion alignment, the accelerometer measurement can be approximated as
\begin{equation}
\boldsymbol{f}^b(t) \approx -\mathbf{C}^b_n(t)\boldsymbol{g}^n(t),
\label{eq:spec_force}
\end{equation}
where $\boldsymbol{g}^n(t)$ is the local gravity vector.

Substituting \eqref{eq:att_decomposition} into \eqref{eq:spec_force} yields
\begin{equation}
\mathbf{C}^{b_0}_b(t)\boldsymbol{f}^b(t)
\approx
-\mathbf{C}^{b_0}_{n_0}\mathbf{C}^{n_0}_n(t)\boldsymbol{g}^n(t),
\label{eq:adia_basic}
\end{equation}
which can be written in the generic vector observation form
\begin{equation}
\boldsymbol{u}^{b_0}(t) = \mathbf{C}^{b_0}_{n_0}\boldsymbol{u}^{n_0}(t).
\label{eq:adia_vectors}
\end{equation}
Equation \eqref{eq:adia_vectors} is the basis for ADIA.

\subsection{Dual-Vector ADIA (DVA)}
\label{sec:DVA}
When only two non-collinear observation vectors $\boldsymbol{u}_1$ and $\boldsymbol{u}_2$ are available, the constant attitude matrix $\mathbf{C}^{n_0}_{b_0}$ can be obtained analytically as \cite{black_1964_aiaa,Silva_2016}
\begin{equation}
\mathbf{C}^{n_0}_{b_0}
=
\begin{bmatrix}
\left(\boldsymbol{u}_{1}^{n_0}\right)^\top \\
\left(\boldsymbol{u}_{2}^{n_0}\right)^\top \\
\left(\boldsymbol{u}_{1}^{n_0} \times \boldsymbol{u}_{2}^{n_0}\right)^\top
\end{bmatrix}^{-1}
\begin{bmatrix}
\left(\boldsymbol{u}_{1}^{b_0}\right)^\top \\
\left(\boldsymbol{u}_{2}^{b_0}\right)^\top \\
\left(\boldsymbol{u}_{1}^{b_0} \times \boldsymbol{u}_{2}^{b_0}\right)^\top
\end{bmatrix}.
\label{eq:DVA}
\end{equation}
The latter version of ADIA is frequently referred to as Dual-Vector ADIA (DVA) \cite{frutuoso_2023_oceaneng}.

\subsection{Optimization-Based ADIA (OBA)}\label{sec:OBA}
For OBA, in turn, the initial attitude is estimated by solving a Wahba-type optimization problem formulated in quaternion space \cite{Wahba_1966}. The optimal quaternion $\boldsymbol{q}=[s\;\boldsymbol{\eta}^\top]^\top$ is obtained from \cite{wu_2013}
\begin{equation}
\min_{\boldsymbol{q}} \ \boldsymbol{q}^\top \mathbf{K}\boldsymbol{q},
\quad
\text{s.t.}\quad \mathbf{q}^\top\mathbf{q}=1,
\label{eq:oba_quat}
\end{equation}
where the matrix $\mathbf{K}$ is defined as
\begin{multline}
\mathbf{K}
=
\sum_{k=1}^{N}
\left(
\mathbf{H}^+(\boldsymbol{u}^{n_0}(t_k))
-
\mathbf{H}^-(\boldsymbol{u}^{b_0}(t_k))
\right)^\top \\
\left(
\mathbf{H}^+(\boldsymbol{u}^{n_0}(t_k))
-
\mathbf{H}^-(\boldsymbol{u}^{b_0}(t_k))
\right),
\end{multline}
with
\begin{align}
\mathbf{H}^+(\boldsymbol{u}) &=
\begin{bmatrix}
0 & -\boldsymbol{u}^\top \\
\boldsymbol{u} & [\boldsymbol{u}\times]
\end{bmatrix},\\
\mathbf{H}^-(\boldsymbol{u}) &=
\begin{bmatrix}
0 & -\boldsymbol{u}^\top \\
\boldsymbol{u} & -[\boldsymbol{u}\times]
\end{bmatrix}.
\end{align}
The optimal quaternion corresponds to the eigenvector associated with the smallest eigenvalue of $\mathbf{K}$, and the initial attitude matrix is recovered as \cite{Wu_2011}
\begin{equation}
\mathbf{C}^{n_0}_{b_0} = \mathbf{C}(\mathbf{q}) = \mathbf{C}^{b_0}_{n_0}{}^\top.
\end{equation}

\subsection{Observation Vectors}\label{sec:obsrv_vec}
For unaided, i.e., self-alignment, the observation vectors are computed as \cite{frutuoso_2025}
\begin{equation}
\left\{
\begin{aligned}
\boldsymbol{u}^{b_0}(t) &= -\int_{t_0}^{t} \mathbf{C}^{b_0}_b(\tau)\boldsymbol{f}^b(\tau)\,d\tau,\\
\boldsymbol{u}^{n_0}(t) &= \int_{t_0}^{t} \mathbf{C}^{n_0}_n(\tau)\boldsymbol{g}^n(\tau)\,d\tau,
\end{aligned}
\right.
\label{eq:I-DVA}
\end{equation}
or, in instantaneous form,
\begin{equation}
\left\{
\begin{aligned}
\boldsymbol{u}^{b_0}(t) &= -\mathbf{C}^{b_0}_b(t)\boldsymbol{f}^b(t),\\
\boldsymbol{u}^{n_0}(t) &= \mathbf{C}^{n_0}_n(t)\boldsymbol{g}^n(t).
\end{aligned}
\right.
\label{eq:A-DVA}
\end{equation}

When employing \eqref{eq:I-DVA}, DVA and OBA are respectively referred to as I-DVA and I-OBA \cite{frutuoso_2023_oceaneng}. When using \eqref{eq:A-DVA} instead, they are called A-DVA and A-OBA, also respectively.
\section{Proposed Approach}\label{prop_approach}
\begin{figure}
    \centering
    \includegraphics[width=0.95\linewidth]{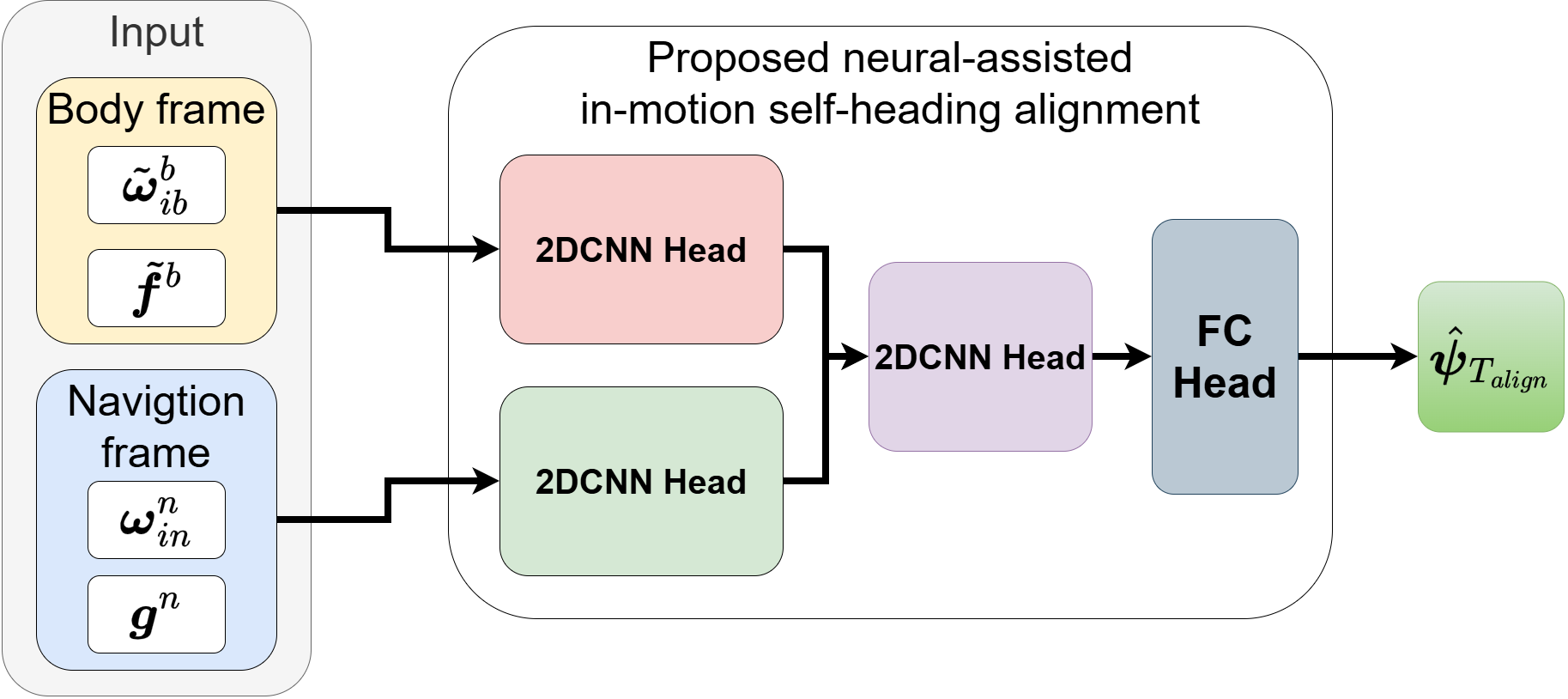}
    \caption{Block diagram showing our proposed data-driven approach for self-heading estimation.}
    \label{fig_general_prop_app_diag}
\end{figure}

\noindent In this section, we present HeadingNet, our proposed, model-free, neural-assisted in-motion self-heading alignment. The proposed approach estimates the heading angle of the vessel at the end of the alignment time based on four types of inputs:  1) the measured angular velocity, $\tilde{\boldsymbol{\omega}}_{ib}^{b}$, 2) the measured specific force, $\tilde{\boldsymbol{f}}^{b}$, 3) the transport rate, $\boldsymbol{\omega}_{in}^{n}$, and 4) the gravity vector, $\boldsymbol{g}^{n}$. The motivation for using these four inputs is twofold and follows the model-based approaches for self-heading alignment. First, all four inputs are used as in the model-based approaches. Second, the inputs are given in two different reference frames, enabling the theoretical foundation for solving the initial heading angle. Additionally, as in the model-based approaches, the same sensors are used. Besides the inertial sensors, the global navigation satellite system (GNSS) provides positioning, allowing for the calculation of the transport rate and gravity vector. Finally, as in any self-heading alignment at sea, we assume quasi-static conditions such that the specific force and angular velocity vectors approximately measure the gravity and the transport rate, respectively.\\
HeadingNet is a neural network (NN) architecture, consisting of several two dimensional convolutional neural networks (2DCNN), followed by several fully connected (FC) layers. HeadingNet receives the four inputs measured during a predefined alignment time, $T_{Align}$, and outputs the heading estimation at the end of that alignment time, $\hat{\psi}_{T_{Align}}$, as illustrated in Figure \ref{fig_general_prop_app_diag}.
\subsection{HeadingNet Backbone Architecture}\label{backcone_arch_sec}
\begin{figure*}[t]
  \includegraphics[width=\textwidth]{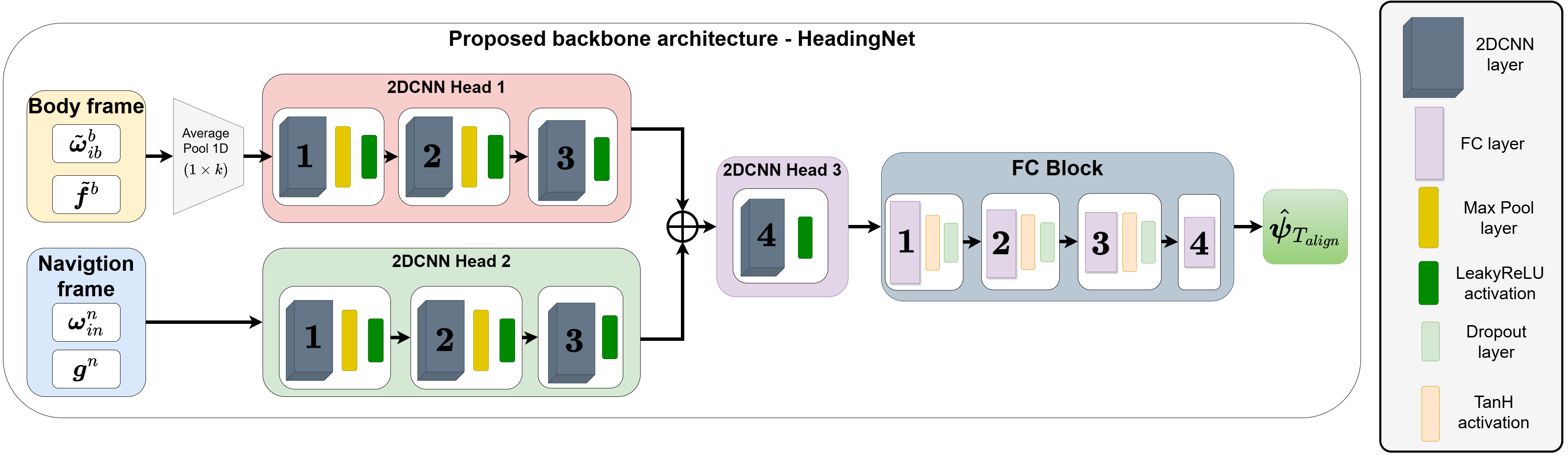}
  \caption{Detailed block diagram showing the proposed neural-assisted in-motion self-heading alignment architecture, that is HeadingNet, and the possible variation due to $T_{Align}$ variations. }\label{fig:full_arch_diagram}
\end{figure*}
\begin{figure}
    \centering
    \includegraphics[width=0.99\linewidth]{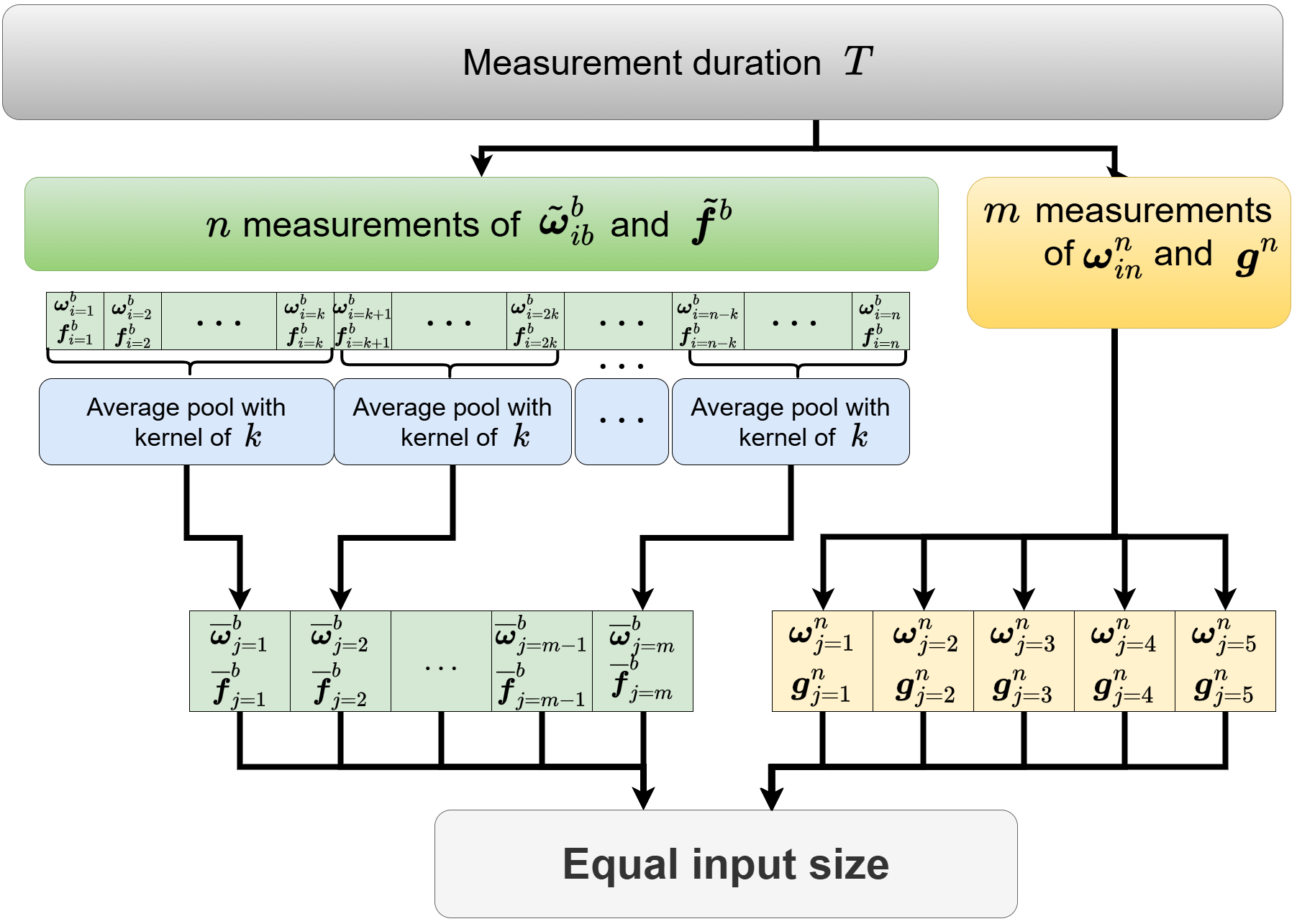}
    \caption{Illustration of an average pooling operation with a kernel of size $k$, applied over a $T$ second measurement window containing $n$ IMU measurements and $m$ transport-rate and gravity measurements, where $n>m$.}
    \label{fig_avg_pool_diag}
\end{figure}

Figure \ref{fig:full_arch_diagram} shows a detailed description of the backbone architecture of HeadingNet, showing the 2DCNN and FC blocks. The four inputs are split by the frame they are measured in, and processed by two separate 2DCNN heads, Head 1 and Head 2. Then, the output is concatenated and processed through an additional 2DCNN head, Head 3. Lastly, it is flattened and passed through a FC block consisting of four FC layers. The 2DCNN heads are utilized for feature and information extraction both in the spatial and temporal domains, with the first two 2DCNN heads used for lower-level feature extraction, and the later Head 3 is utilized for finer feature and information extraction. The FC block is utilized for the final regression and estimation of the heading angle at the end of the alignment time, $\hat{\psi}_{T_{Align}}$. \\
The architecture of Heads 1 and Head 2 is identical, and consists of three 2DCNN layers, where the first two layers are followed by a max pooling layer \cite{nirthika2022pooling} and a leaky rectified linear unit (LeakyReLU) activation function \cite{dubey2022activation,sharma2017activation}, while the third 2DCNN layer is followed only by a LeakyReLU activation function. The first 2DCNN layer has an input of 1 channel and 16 output channels with a kernel sized $(2 \times T_{Align})$, the second 2DCNN layer has 16 input channels and outputs 32 channels, and has a kernel size $(2 \times 0.75 \cdot T_{Align})$, the third 2DCNN layer has 32 input channels and 64 output channels with a kernel sized $(2 \times 0.5 \cdot T_{Align})$. The two max pooling layers which are applied after the first and second 2DCNN layers have the same kernel size of $(1 \times 2)$. The LeakyReLU activation function has a scaling factor for the negative values of $\alpha_{T_{align}}$, and has the same value for all LeakyReLU activation functions used in the HeadingNet architecture.\\
The output of Head 1 and Head 2 are concatenated and processed by an additional 2DCNN head (Head 3), for further feature and information extraction. In Head 3, the concatenated input is processed by a fourth 2DCNN layer, with 64 input and 128 output channels, and a kernel size $(3 \times 3)$. This layer is followed by a LeakyReLU activation function.\\
The output of Head 3 is encoded in a latent representation and has several dimensions since it was previously processed by 2DCNN layers for feature, hidden dependencies, and correlations extraction. This output is flattened and passed to the FC block consisting of four FC layers. In the first three, each FC layer is followed by a hyperbolic tangent (TanH) nonlinear activation function applied to prevent the NN from acting as a large linear regressor model \cite{sharma2017activation}. After the TanH, function a Dropout layer is applied for regularization \cite{srivastava2014dropout}, while the fourth and last FC layer directly outputs the estimated heading angle for the alignment time $\hat{\psi}_{T_{Align}}$. \\
The first FC layer input dimension is $h_{in}^{1}$ and the output dimension is 512, the second FC layer has an input size of 512 and output size of 128, the third FC layer has an input size of 128 and an output size of 32, while the fourth and last FC layer has an input size of 32 and output size of one, that is the heading estimation $\hat{\psi}_{T_{Align}}$. In all three dropout layers, the probability parameter of $p_{T_{Align}}$ is applied for all three.\\
As presented in Figure \ref{fig:full_arch_diagram}, an average pooling layer is applied only on two out of the four inputs, namely the angular velocity, $\tilde{\boldsymbol{\omega}}_{ib}^{b}$, and the specific force, $\tilde{\boldsymbol{f}}^{b}$, vectors. The average pooling layer \cite{nirthika2022pooling} is applied to tackle the temporal size difference between the angular velocity and specific force, and the gravity and transport rate, thus equating their sizes before entering Head 1. The average pooling layer consists of six input channels, each one with a kernel of $(1\times k)$ and a stride of $k$, thus reducing the temporal dimension of the input by a constant factor of $k$. Figure \ref{fig_avg_pool_diag} shows a general example of the average pooling layer receiving a measurement of duration $T$ seconds. This measurement holds $n$ specific force and angular velocity measurements, and $m$ transport rate and gravity measurements, where $n > m$. By applying the average pool layer with a kernel of $k$, we reduce the $n$ measurements by a factor of $k$, by averaging every $k$ non-overlapping measurements, resulting in $m$ specific force and angular velocity measurements. Note that the average pooling has no weights which are optimized during the training process, hence is not considered a learnable layer.
\subsection{Neural network Formulation and Training Process}\label{prop_nn_math}
\noindent In this section the mathematical formulation and notation for all the neural network components and layers are described.
\subsubsection{Fully Connected Neural Network}\label{FC_subsec}
\noindent In a fully connected layer, each neuron in the current layer is connected to each neuron in the preceding layer and to each neuron in the following layer, where each neuron in the FC layer is defined by \cite{prince2023understanding,suzuki2013artificial,sze2017efficient}:
\begin{equation}\label{neuron_def}
    {z}_{i}^{(l)} = \sum_{j=1}^{n_{(l-1)}} {{\omega}_{ij}^{(l)}{a}_{j}^{(l-1)}} + {b}_{i}^{(l)}
\end{equation}
where ${\omega}_{ij}^{(l)}$ is the weight and ${b}_{i}^{(l)}$ is the bias of the $i^{th}$ neuron in the $l^{th}$ layer, $a_{j}^{l-1}$ is the output of the $j$ neuron from the $(l-1)^{th}$ layer and is connected to the said $i^th$ neuron that is connected to the $j^{th}$ neuron in the $(l-1)^{th}$ layer.
\subsubsection{Two Dimensional Convolution Neural Network}\label{cnn_subsec}
Each CNN layer is comprised of several $m_{1} \times m_{2}$ matrices, where each entry in the matrix is a neuron and the matrix is called a kernel. A CNN kernel is defined as: \cite{prince2023understanding,sze2017efficient}:
\begin{equation}\label{conv_kernel_eq_def}
    C_{ij}^{(l)} = \sum_{\alpha=0}^{m_1} \sum_{\beta=0}^{m_2} {{\omega}_{\alpha\beta}^{(r)}{a}_{(i+\alpha)(i + \beta)}^{(l-1)}} + {b}^{(r)}
\end{equation}
where $\omega^{r}_{\alpha \beta}$ is the weight at position $(\alpha,\beta)$ of the kernel at the $r^{th}$ layer, $a$ is the output of the previous layer $(l-1)$, and for each CNN kernel a single bias is added for the $r^{th}$ layer, denoted as $b^{(r)}$. The term $C_{ij}^{(l)}$ is the CNN kernel output.
\subsubsection{Nonlinear Activation Functions}\label{activation_func_subsec}
In this section, we cover two nonlinear activation functions which are applied in our proposed HeadingNet architecture. Activation functions are applied in deep learning models to improve the NN performance and to prevent the NN model from acting as a large linear regressor \cite{apicella2021survey,sharma2017activation,dubey2022activation}. Traditionally, activation function are applied as follows \cite{sharma2017activation}:
\begin{equation}\label{general_act_func_eq}
    a_{i}^{(l)} = h({z}_{i}^{(l)})
\end{equation}
where ${z}_{i}^{(l)}$ is the output we apply the activation functions $h$ to, and $a_{i}^{(l)}$ is the output. Notice that activation function is applied to the same neuron denoted by a subscript of $i$ in the $l^{th}$ layer. We employ two activation functions:
\begin{enumerate}
    \item \textbf{Leaky rectified linear unit (LeakyReLU)}: an activation function which does not change the input for positive values while applying a scale, or a coefficient, to negative values, as follows \cite{sharma2017activation}:
    \begin{equation}\label{leakyrelu}
        LeakyReLU(C_{ij}^{(l)}x) = 
        \begin{cases}
                C_{ij}^{(l)},& \text{if } C_{ij}^{(l)} \geq 0\\
                \alpha \cdot C_{ij}^{(l)},              & \text{otherwise}
                \end{cases}
    \end{equation}
    where $C_{ij}^{(l)}$ is the LeakyReLU input, and $\alpha$ is the negative values coefficient.
    \item \textbf{Hyperbolic Tangent Function (Tanh)}: A nonlinear activation function which applies the following to an input \cite{dubey2022activation}:
    \begin{equation}\label{tanh_eq}
        Tanh({z}_{i}^{(l)}) = \frac{e^{{z}_{i}^{(l)}} - e^{-{z}_{i}^{(l)}}}{e^{{z}_{i}^{(l)}} + e^{-{z}_{i}^{(l)}}}
    \end{equation}
    where ${z}_{i}^{(l)}$ is the input to the TanH activation function and $e$ is the Euler constant.
\end{enumerate}

\subsubsection{Regularization}\label{regularization_subsec}
Overfitting is a term describing a phenomenon where a given NN is trained on a training set and achieves a low error om the training set by learning noise patterns unique to the set, however, failing to generalize to unseen data and performing poorly in evaluation \cite{salman2019overfitting}. In this work, to mitigate overfitting and improve the generalization of the NN, we utilize dropout regularization \cite{srivastava2014dropout}. By applying dropout, a neuron is activated during training with probability $1-p$, as follows \cite{srivastava2014dropout}:
\begin{equation}
\begin{split}
    r_{i}^{l} & \sim Bermoulli(p) \\
    \Tilde{a}_{i}^{l} & = r_{i}^{l} \cdot a_{i}^{l} \\
    a_{j}^{l+1} & = {{\omega}_{ji}^{(l+1)}\Tilde{a}_{i}^{(l)}} + {b}_{j}^{(l+1)}
\end{split}
\end{equation}
where $r_{i}^{l}$ is a random variable that is sampled from a $Bernoulli$ distribution with a probability of $p$, $a_{i}^{l}$ is the input, which is the activation function output of the $i^{th}$ neuron of the previous $l^{th}$ layer \eqref{neuron_def}, and $\cdot$ denotes the element-wise multiplication of $r^{l}_{i}$ with $a_{i}^{l}$, which results in the dropout output $\Tilde{a}_{i}^{l}$. Once dropout was applied, the output, $\Tilde{a}_{i}^{l}$, is passed to the next layer connected neurons which are the weights ,${\omega}_{ji}^{(l+1)}$ and biases, ${b}_{j}^{(l+1)}$, of the next $(l+a)$  layer. Notice that in our proposed approach the output of the activation function $a_{i}^{l}$ applied is the TanH activation function \eqref{tanh_eq}, since dropout is utilized in the FC blocks as described in Section \ref{backcone_arch_sec}.

\subsubsection{Training Process}\label{training_process_loss_subsec}
\noindent During training, the goal is to determine the best weights and biases of all the neurons in the NN in order to achieve the lowest loss. The loss value is calculated by utilizing some loss function which compares the difference between the true value and the estimated value. \\
In this work, the estimated value is an angle with a cyclic nature, unlike the unidirectional scale of traditional loss functions. Therefore, to effectively minimize the error between two angles, we adopt the cyclic loss mean square error (CMSE) as proposed in \cite{ENGELSMAN2026112842}. To this end, let:
\begin{equation}\label{lcme_delta_psi_eq}
\Delta \boldsymbol{\psi}_{i} = \hat{\boldsymbol{\psi}}_i - \boldsymbol{\psi}_i \: \: , \forall i \in {1, \ldots,N}
\end{equation}
where $\hat{\boldsymbol{\psi}}_i$ is the predicted heading angle, $\boldsymbol{\psi}_i$ is the true heading angle, and $N$ is the number of samples in the batch. Then we calculate the error in a quadrant aware function, the atan2:
\begin{equation}\label{lcme_err_def_eq}
\boldsymbol{err}_i = \mathrm{atan2}\big(\sin(\Delta \boldsymbol{\psi}_{i}), \cos(\Delta \boldsymbol{\psi}_{i})\big) \: , \forall i \in {1, \ldots,N}
\end{equation}
Then, the CMSE is defined by:
\begin{equation}\label{final_cmse_loss_eq}
\mathcal{L}_{\mathrm{CMSE}} = \big(\frac{1}{N}\sum_{i = 1}^{N} \boldsymbol{err}_i^2 \big) \cdot \lambda
\end{equation}
where $\lambda$ is a scaling factor which multiplies the CMSE loss. To estimate the optimal numeric values of weights and biases given the derivatives a gradient decent algorithm is applied: 
\begin{equation}\label{grad_desc_eq}
    \boldsymbol{\theta} = \boldsymbol{\theta} - \eta\nabla_{\theta}\mathcal{L}_{\mathrm{CMSE}}(\theta), \: \:\:\: \boldsymbol{\theta} = [\omega \: \: \: b]^T
\end{equation}
where $\nabla_{\theta}$ is the gradient with respect to the weights and biases, and $\eta$ is the learning rate which is a hyperparameter that is set at the start of training.\\
To estimate the correct gradient step, the adaptive moment estimation with weight decay (AdamW) is used \cite{loshchilov2017decoupled}. AdamW offers to decouple the weight decay regularization from the optimization step performed by the optimizer with respect to the loss function.\\
To further improve the accuracy of the heading estimation and improve convergence \cite{He_2019_CVPR, loshchilov2017decoupled} a learning rate scheduler was applied, the StepLR \cite{steplr_pytorch}. The main idea is that, to improve results during training, every few predefined steps, that is every predefined number of epochs, the learning rate decays by a predefined factor denoted as $\gamma_{LR}$.
\section{HeadingNet Architecture Variations}\label{headingnet_variations_sec}
In real world scenarios the allowed alignment time, $T_{Align}$, varies due to practical considerations. In Section \ref{prop_approach}, we provided a general framework for our HeadingNet architecture. Since the input size increases with alignment time, the backbone NN architecture is adapted to accommodate the larger input. In particular, the increased input temporal dimension results in larger intermediate latent representations produced by the 2DCNN layers and other NN components
throughout the network. Therefore, several modifications are
introduced to the backbone architecture. These include two main modifications: The first is the addition of a max pooling layer after the third 2DCNN layer in Head 1 and Head 2. The second is the addition of a fifth 2DCNN layer followed by a LeakyReLU activation in Head 3.\\
In this work, we consider and evaluate five different alignment durations of $10,30,60,90,$ and $120$ seconds. While $10,30,$ and $60$ seconds represent rapid alignment durations, the model-based approaches require $120$ seconds to reach sufficient performance. The following describes the modifications made as a function of the alignment time:
\begin{enumerate}
    \item \textbf{HeadingNet10}: This alignment time architecture is identical to the backbone NN architecture presented in Section \ref{backcone_arch_sec}.
    \item \textbf{HeadingNet30}: Compared to the baseline, all 2DCNN layers in Head 1 and Head 2, as well as the fourth 2DCNN layer in Head 3, have different kernel sizes. In addition, we add a fifth 2DCNN layer and a LeakyReLU activation in Head 3.
    \item \textbf{HeadingNet60}: Compared to the backbone, the 2DCNN layer kernels in Head 1, Head 2, and Head 3 have different sizes. Additionally, in Head 3, a fifth 2DCNN layer followed by a LeakyReLU activation is added. Finally, the input size of the first FC layer is modified.
    \item \textbf{HeadingNet90}: Compared to the backbone, the 2DCNN layer kernel sizes in Head 1, Head 2, and Head 3 are modified. In addition, a max pooling layer is inserted between the third 2DCNN layer and the subsequent LeakyReLU activation in Head 1 and Head 2. Finally, in Head 3, a fifth 2DCNN layer followed by a LeakyReLU activation is added.
    \item \textbf{HeadingNet120}: Compared to the baseline, five modifications are made. First, the 2DCNN kernel sizes in Head 1, Head 2, and Head 3 differ. Second, a max pooling layer is inserted between the third 2DCNN layer and the subsequent LeakyReLU activation in Head 1 and Head 2. Third, a fifth 2DCNN layer followed by a LeakyReLU activation is added in Head 3. Fourth, the input size of the first FC layer, $h^{1}_{in}$, is varied. Finally, the dropout probability in the FC block is modified.
\end{enumerate}
Table \ref{2dcnn_heads_params_tbl} summarizes the main differences for each HeadingNet variation.
\begin{table}[!h]
\centering
\begin{adjustbox}{width = \columnwidth}
\begin{tabular}{|c|c|c|c|c|}
\hline
\begin{tabular}[c]{@{}c@{}}NN Variation/\\ parameter\end{tabular} & \begin{tabular}[c]{@{}c@{}}2DCNN layer\\ number\end{tabular} & \begin{tabular}[c]{@{}c@{}}Kernel\\ size\end{tabular} & \begin{tabular}[c]{@{}c@{}}Max\\ pool\end{tabular} & \begin{tabular}[c]{@{}c@{}}LeakyReLU\\ lambda\end{tabular} \\ \hline
\multirow{5}{*}{\begin{tabular}[c]{@{}c@{}}HeadingNet10\\(backbone architecture)\end{tabular}}                                    & $1^{st}$ Layer                                                    & (2x10)                                                & (1x2)                                              & \multirow{5}{*}{0.05}                                      \\ \cline{2-4}
                                                                  & $2^{nd}$ Layer                                                    & $(2\times7)$                                                 & $(1\times2)$                                              &                                                            \\ \cline{2-4}
                                                                  & $3^{rd}$ Layer                                                    & $(2\times 5) $                                                & None                                               &                                                            \\ \cline{2-4}
                                                                  & $4^{th}$ Layer                                                      & $(3\times 3)$                                                 & \multirow{2}{*}{None}                              &                                                            \\ \cline{2-3}
                                                                  & $5^{th}$ Layer                                                      & None                                                  &                                                    &                                                            \\ \hline
\multirow{5}{*}{HeadingNet30}                                     & $1^{st}$ Layer                                                    & $(2\times 30) $                                               & $(1\times 2)$                                              & \multirow{5}{*}{0.05}                                      \\ \cline{2-4}
                                                                  & $2^{nd}$ Layer                                                    & $(2\times 22) $                                               & $(1\times 2)$                                              &                                                            \\ \cline{2-4}
                                                                  & $3^{rd}$ Layer                                                    & $(2\times 15) $                                               & None                                               &                                                            \\ \cline{2-4}
                                                                  & $4^{th}$ Layer                                                      & $(2\times 3) $                                                & \multirow{2}{*}{None}                              &                                                            \\ \cline{2-3}
                                                                  & $5^{th}$ Layer                                                      & $(2\times 3)$                                                 &                                                    &                                                            \\ \hline
\multirow{5}{*}{HeadingNet60}                                     & $1^{st}$ Layer                                                    & $(2\times 60) $                                               & $(1\times 2)$                                              & \multirow{5}{*}{0.05}                                      \\ \cline{2-4}
                                                                  & $2^{nd}$ Layer                                                    & $(2\times 45) $                                               & $(1\times 2) $                                             &                                                            \\ \cline{2-4}
                                                                  & $3^{rd}$ Layer                                                   & $(2\times 30)$                                                & None                                               &                                                            \\ \cline{2-4}
                                                                  & $4^{th}$ Layer                                                      & $(2\times 6)$                                                 & \multirow{2}{*}{None}                              &                                                            \\ \cline{2-3}
                                                                  & $5^{th}$ Layer                                                     & $(2\times 3)$                                                 &                                                    &                                                            \\ \hline
\multirow{5}{*}{HeadingNet90}                                     & $1^{st}$ Layer                                                    & $(2\times 90) $                                               & $(1\times 2)$                                              & \multirow{5}{*}{0.1}                                       \\ \cline{2-4}
                                                                  & $2^{nd}$ Layer                                                    & $(2\times 67) $                                               &$ (1\times 2)$                                              &                                                            \\ \cline{2-4}
                                                                  & $3^{rd}$ Layer                                                   & $(2\times 45) $                                               & $(1\times 2)$                                              &                                                            \\ \cline{2-4}
                                                                  & $4^{th}$ Layer                                                     & $(2\times 4) $                                                & \multirow{2}{*}{None}                              &                                                            \\ \cline{2-3}
                                                                  & $5^{th}$ Layer                                                      & $(2\times 3)$                                                 &                                                    &                                                            \\ \hline
\multirow{5}{*}{HeadingNet120}                                    & $1^{st}$ Layer                                                   & $(2\times 120)$                                               & $(1\times 2)$                                              & \multirow{5}{*}{0.05}                                      \\ \cline{2-4}
                                                                  & $2^{nd}$ Layer                                                    & $(2\times 90)$                                                & $(1\times 2)$                                              &                                                            \\ \cline{2-4}
                                                                  & $3^{rd}$ Layer                                                   & $(2\times 60)$                                                & $(1\times 2)$                                              &                                                            \\ \cline{2-4}
                                                                  & $4^{th}$ Layer                                                      & $(2\times 5)$                                                 & \multirow{2}{*}{None}                              &                                                            \\ \cline{2-3}
                                                                  & $5^{th}$ Layer                                                     & $(2\times 3)$                                                 &                                                    &                                                            \\ \hline
\end{tabular}
\end{adjustbox}\caption{Main network parameters including the 2D kernel sizes, the max pool layers kernel sizes, and the LeakyReLU parameters for all five variations of HeadingNet.}\label{2dcnn_heads_params_tbl}
\end{table}\\
Table \ref{variation_and_train_hyperparams_tbl} summarizes the regularization and training hyperparameters, as well as the modifications made to the FC block components for each HeadingNet variation. These regularization and training hyperparameters include the number of epochs, the loss amplification scalar, $\lambda$, and the scheduler and optimizer parameters. The batch size and the scheduler learning-rate decay parameter, $\gamma_{LR}$, remain fixed at 512 and 0.8, respectively, for all HeadingNet variations. The FC block hyperparameters include the input dimension of the first FC layer, $h_{in}^{1}$, and the dropout probability, $p_{T_{Align}}$.\\
As shown in Table \ref{variation_and_train_hyperparams_tbl}, several hyperparameters differ between HeadingNet variations. These differences arise from two main factors. First, the network weights are initialized with different random values at the beginning of each training process. Second, as described in Section \ref{data_sub_sec}, the dataset is processed and shuffled differently for each variation. Due to these sources of stochasticity in training, the optimization process may converge to different minima of the loss function, potentially resulting in sub optimal heading estimation. Therefore, the hyperparameters are tuned separately for each HeadingNet variation to achieve optimal heading estimation performance.
\begin{table}[!h]
\begin{adjustbox}{width = \columnwidth}
\begin{tabular}{|c|c|c|c|c|c|c|c|c|}
\hline
             \begin{tabular}[c]{@{}c@{}}HeadingNet\\variation\end{tabular} & $h^{1}_{in}$ & $p_{T_{Align}}$ 
              & Epochs & $\lambda$ & \begin{tabular}[c]{@{}c@{}}Learning\\ rate\end{tabular} & \begin{tabular}[c]{@{}c@{}}Weight\\ decay\end{tabular} & \begin{tabular}[c]{@{}c@{}}Scheduler\\ step\end{tabular} \\ \hline
HeadingNet10  & 512 & 0.2 & 1000                                                     & 10    & 0.0009                                                  & 0.08                                                   & 120                                                                                                           \\ \hline
HeadingNet30 & 512 & 0.2 & 1000                                                     & 10    & 0.0008                                                  & 0.08                                                   & 120                                                                                                            \\ \hline
HeadingNet60 & 1024 & 0.2 & 400                                                      & 10    & 0.0008                                                  & 0.08                                                   & 80                                                                                                             \\ \hline
HeadingNet90 & 512 & 0.2 & 500                                                      & 100   & 0.0005                                                  & 0.8                                                    & 150                                                                                                             \\ \hline
HeadingNet120 & 1024 & 0.3 & 300                                                      & 10    & 0.0006                                                  & 0.08                                                   & 50                                                                                                             \\ \hline
\end{tabular}
\end{adjustbox}\caption{The training hyperparameters and the FC block parameters used in the training of the HeadinNet variation.}\label{variation_and_train_hyperparams_tbl}
\end{table}
\section{Experimental Results}\label{res_sec}
This section provides a description of the dataset used to evaluate the proposed approach, the data processing pipeline used to create the training and evaluation sets for HeadingNet, and lastly the results.
\subsection{Data Collection and Processing}\label{data_sub_sec}
\begin{figure}[h]
\centering
  \includegraphics[width=0.85\columnwidth]{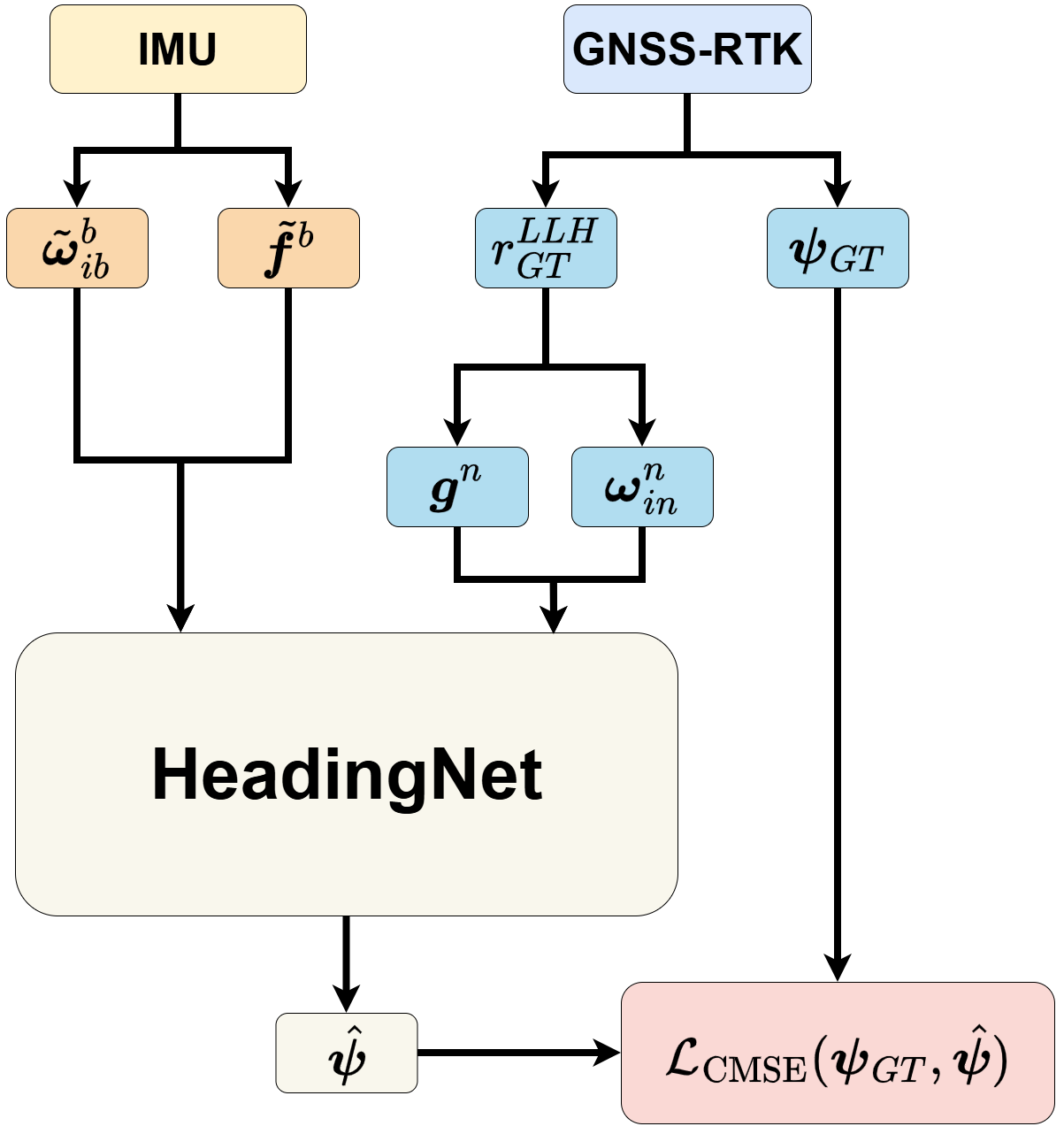}
  \caption{Data flow diagram of starting from the raw measurements from the ASV sensors, the IMU and GNSS-RTK, to our proposed approach HeadingNet.}\label{fig:asv_sensors_data_flow}
\end{figure}

\begin{figure}[h]
\centering
  \includegraphics[width=0.95\columnwidth]{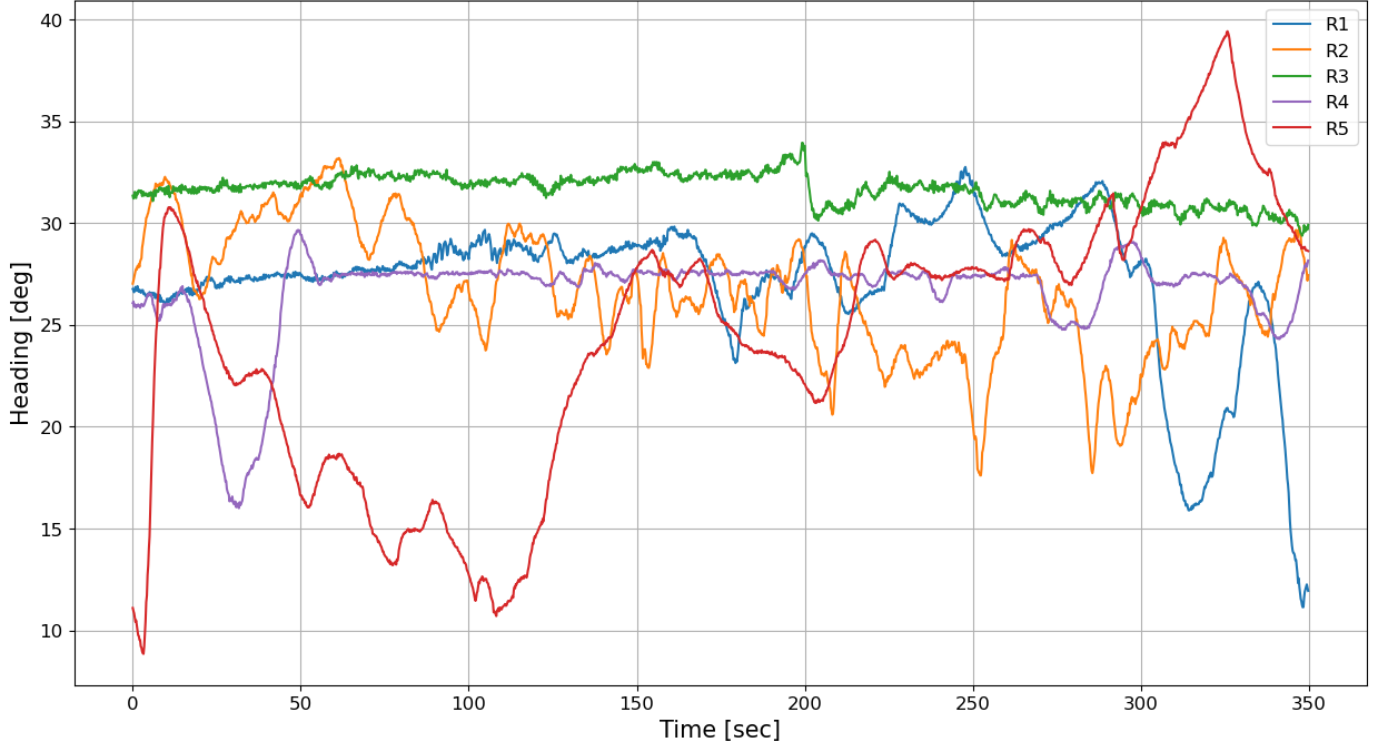}
  \caption{Heading angles plot of the first $350$ seconds for the five recorded trajectories, R1 to R5.}\label{fig:headings_plot_350sec}
\end{figure}

\begin{figure}[h]
\centering
  \includegraphics[width=0.95\columnwidth]{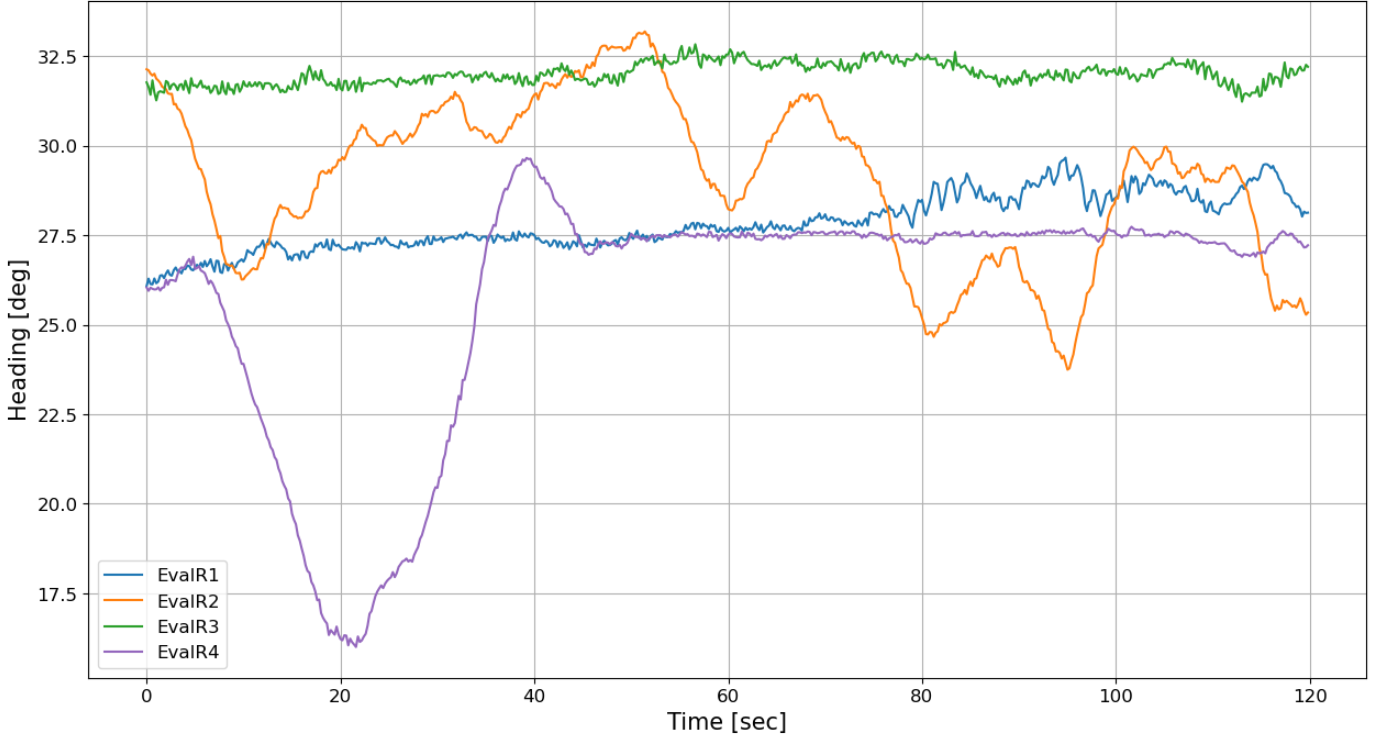}
  \caption{Heading angles plot of the four evaluation recordings, EvalR1 to EvalR4.}\label{fig:eval_recordings_heading_plot}
\end{figure}

To train and validate HeadingNet, real-world data were recorded using an ASV tied to a pier \cite{frutuoso_2023_oceaneng}. Data were collected over five different days under varying sea conditions, resulting in five recordings of the ASV in mooring conditions, reflecting diverse real-world conditions. The ASV was equipped with an IMU and a dual GNSS receiver with real-time kinematics (GNSS-RTK) antennas, providing accurate heading estimation and precise positioning. The GNSS-RTK receiver provides a position accuracy of $0.8$ cm and a heading accuracy of $0.09$ degrees. Table \ref{imu_specs_tbl} lists the specifications of the onboard IMU used in the experiments. Additionally, the IMU sampling rate was $100$ Hz, while the GNSS-RTK sampling rate was $5$ Hz.
\begin{table}[h]
\centering
\begin{adjustbox}{width = 0.8\columnwidth}
\begin{tabular}{|c|c|c|}
\hline
              & \begin{tabular}[c]{@{}c@{}}Bias in-run\\ stability\end{tabular} & \begin{tabular}[c]{@{}c@{}}Angular/velocity\\ random walk\end{tabular} \\ \hline
Gyroscope     & 0.02 $[\degree /s]$                                         & 0.032 $[\degree/\sqrt{hr}]$                                               \\ \hline
Accelerometer & 1000 $[\micro g]$                                            & 0.012 $[m/s/\sqrt{hr}]$                                               \\ \hline
\end{tabular}
\end{adjustbox}\caption{Gyroscope and accelerometer specifications mounted on the ASV used in the  experiments.}\label{imu_specs_tbl}
\end{table}
\noindent The latitude obtained from the position is  used to compute the transport rate, $\boldsymbol{\omega}_{in}^{n}$, and the gravity, $\boldsymbol{g}^{n}$, which constitute two of the four inputs to HeadingNet. The IMU provides the remaining two inputs: the specific force, $\tilde{\boldsymbol{f}}^{b}$ and angular velocity measurements, $\tilde{\boldsymbol{\omega}}_{ib}^{b}$. Note that the IMU and GNSS-RTK operate at different sampling rates, resulting in 20 times as many IMU measurements as GNSS-RTK measurements. To address this difference, average pooling is applied within HeadingNet, as described in Figure \ref{fig:full_arch_diagram}. In addition, since the GNSS-RTK provides highly accurate heading measurements, they are used as ground truth (GT) heading labels during training, denoted as $\boldsymbol{\psi}_{GT}$. Once the four inputs are obtained from the IMU and GNSS-RTK, they are passed to HeadingNet and the baseline approaches. Figure \ref{fig:asv_sensors_data_flow} illustrates the data flow from the ASV sensors for HeadingNet.
\\
\noindent The five recordings are labeled R1 to R5, with a total duration of $131$ minutes. Recordings R2 and R3 are significantly longer than the others, with durations of $48.5$ and $56.6$ minutes, respectively. The durations of recordings R1, R4, and R5 are $11.4$, $7.1$, and $7.3$ minutes, respectively. Figure \ref{fig:headings_plot_350sec} shows the ASV heading for recordings R1–R5 during the first $350$ seconds of each recording.
By examining the heading plots in Figure \ref{fig:headings_plot_350sec}, it can be observed that the fifth recording, R5, exhibits larger heading variations compared to recordings R1–R4. Therefore, R5 was used only for training and not for validation. Additionally, since the recording durations of the five trajectories vary significantly, particularly for R2 and R3, the dataset was split into evaluation and training sets as follows:
\begin{enumerate}
    \item \textbf{Evaluation set}: A 120 second section from each of the four recordings, R1–R4, was reserved for evaluation. Each section starts at 10 seconds and concludes at 130 seconds. The total duration of the evaluation set is 8 minutes. Figure \ref{fig:eval_recordings_heading_plot} shows the heading angle for the four evaluation recordings, referred to as EvalR1 to EvalR4.
    \item \textbf{Training set}: This set consists of the remaining data from the five recordings, R1–R5, starting at 130 seconds onward. Since recordings R2 and R3 are significantly longer than the others, only data up to 600 seconds from the start of each recording were used. For recordings R1, R4, and R5, the data from 130 seconds until the end of the recording were used. This results in a total training duration of $35.07$ minutes. Table \ref{training_set_times_tbl} provides a detailed description of the training set used from each recording, referred to as TrainR1 to TrainR5.
\end{enumerate}
\begin{table}[h]
\centering
\begin{adjustbox}{width = \columnwidth}
\begin{tabular}{|c|c|c|c|c|c|c|}
\hline
                & TrainR1    & TrianR2   & TrianR3   & TrianR4    & TrianR5    & Total  \\ \hline
Time {[}min{]} & 9.26  & 7.83 & 7.83 & 4.96  & 5.19  & 35.07  \\ \hline
\end{tabular}
\end{adjustbox}\caption{Duration of each trajectory in the training set.}\label{training_set_times_tbl}
\end{table}

\begin{figure*}[!h]
\centering
  \includegraphics[width=0.95\textwidth]{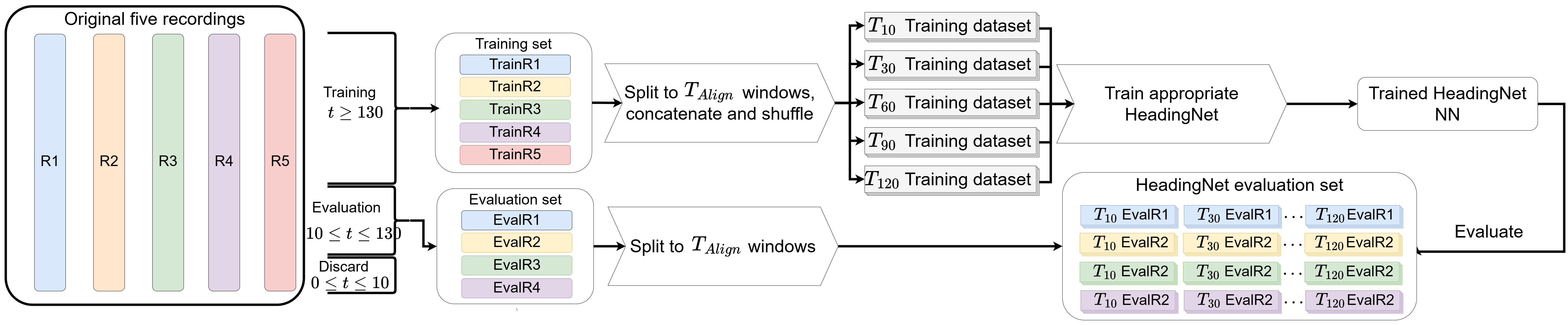}
  \caption{Block diagram illustrating the data preparation pipeline to generate the training sets and datasets and evaluation set to train and evaluation all HeadingNet variations. }\label{fig:data_preparation_pipeline_fig}
\end{figure*}

For each evaluated alignment time, the temporal dimension of the four inputs to HeadingNet was varied, therefore, the training set was processed differently for each alignment time. In the training set, for each alignment time, the five training recordings, TrainR1 to TrainR5, were split into $T_{Align}$ second sliding windows with a stride of one second, after which the windows were concatenated, shuffled, and then divided into batches of 512. As a result, each data point is now $T_{Align}$ seconds long and contains the four inputs to HeadingNet. Each data point is assigned the GT heading label $\psi_{GT}$, defined as the last heading measurement within the corresponding sliding time window. Overall, this process creates five unique training datasets, each corresponding to a different alignment time and its associated HeadingNet variation.\\
Regarding the four recordings in the evaluation set, EvalR1 to EvalR4, the same data processing pipeline was applied with two main differences. First, the stride length was set equal to the window size in order to divide each evaluation recording into non-overlapping windows. Second, the resulting windows were not shuffled, allowing them to remain in their original temporal order. Consequently, for each of the four evaluation recordings, five different evaluation sub-recordings were created, corresponding to the five evaluated alignment times.\\
For the model-based approaches, no training was required, and no splitting of the four evaluation recordings, EvalR1 to EvalR4, was necessary. Each baseline approach estimated the heading angle for the four evaluation recordings at each alignment window, since the baseline methods can estimate the heading at any desired time interval. Figure \ref{fig:data_preparation_pipeline_fig} illustrates the complete data processing and preparation pipeline for the training and evaluation sets used to train and evaluate HeadingNet and its variations.
\subsection{Performance Metric}\label{eval_metrics}
The metric employed to evaluate the heading accuracy is the absolute error (AE) of the heading angle:
\begin{equation}\label{ae_def_eq}
    AE = \mid{\hat{\boldsymbol{\psi}}_{NN} - \boldsymbol{\psi}}\mid
\end{equation}
where $\hat{\boldsymbol{\psi}}_{NN}$ is the heading estimation by HeadingNet, and $\boldsymbol{\psi}$ is the GT heading angle.
\subsection{Results}
\begin{figure*}[h]
  \centering
  \begin{subfigure}[c]{0.48\textwidth}
    \centering
    \includegraphics[width=\linewidth]{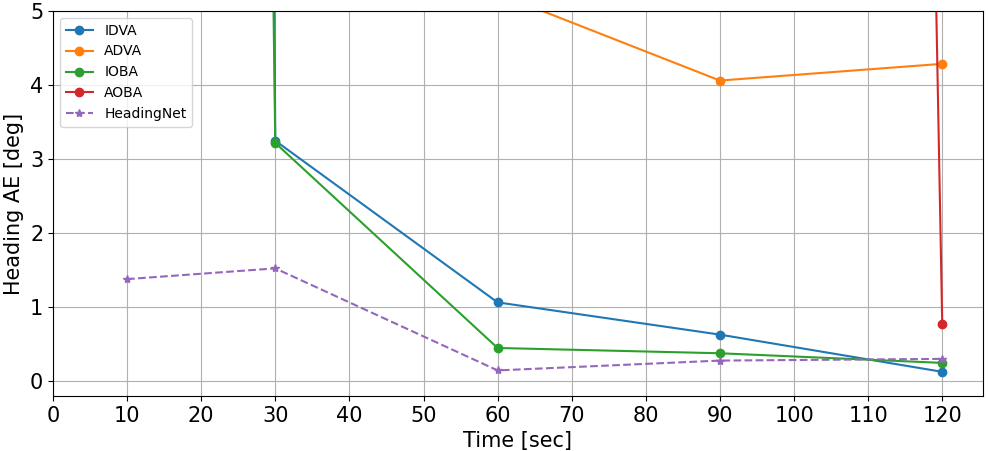}
    \caption{AE results for EvalR1}
    \label{fig:evalr1_ae_res_plot}
  \end{subfigure}
  \hfill
  \begin{subfigure}[c]{0.48\textwidth}
    \centering
    \includegraphics[width=\linewidth]{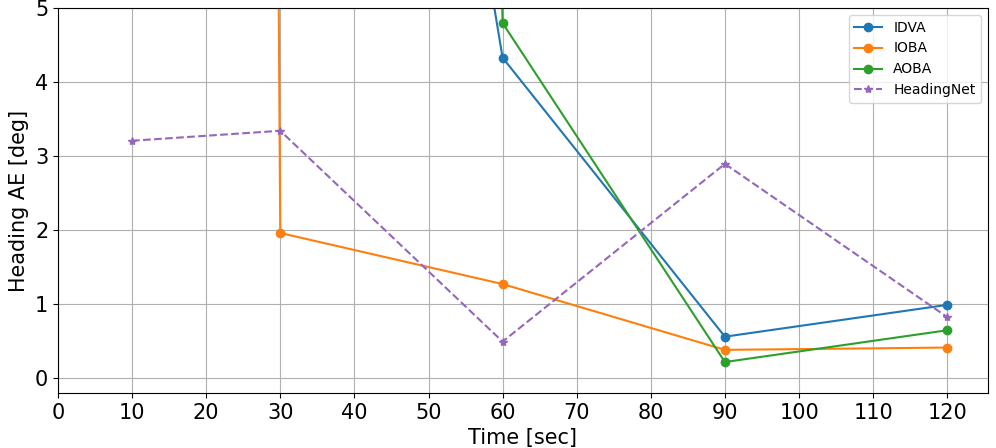}
    \caption{AE results for EvalR2}
    \label{fig:evalr2_ae_res_plot}
  \end{subfigure}
  
  \vspace{0.5cm}
  
  \begin{subfigure}[c]{0.48\textwidth}
    \centering
    \includegraphics[width=\linewidth]{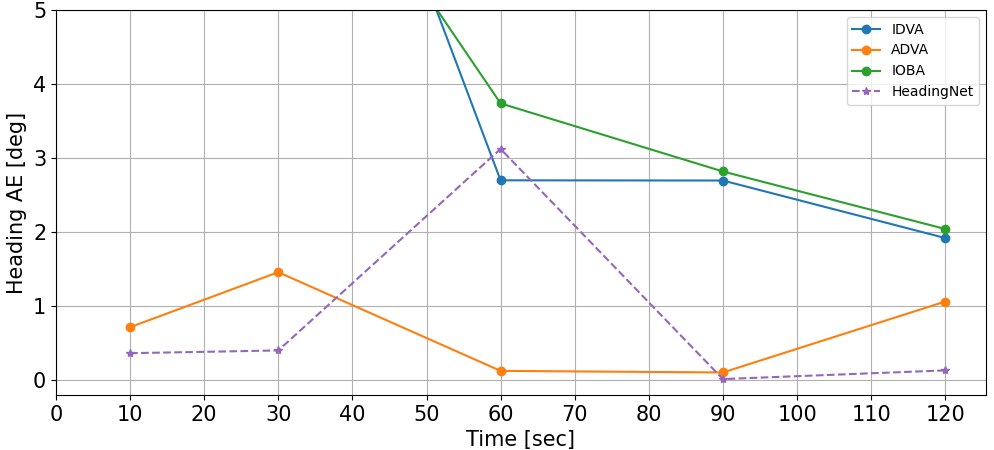}
    \caption{AE results for EvalR3}
    \label{fig:evalr3_ae_res_plot}
  \end{subfigure}
  \hfill
  \begin{subfigure}[c]{0.48\textwidth}
    \centering
    \includegraphics[width=\linewidth]{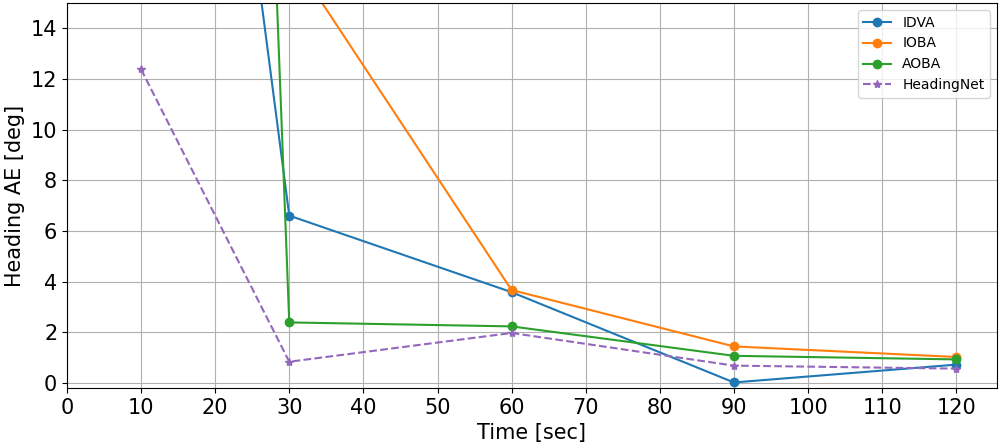}
    \caption{AE results for EvalR4}
    \label{fig:evalr4_ae_res_plot}
  \end{subfigure}
  \caption{AE Results of the heading estimation in degrees for the four evaluation recordings, EvalR1 to EvalR4, at the examined alignment time periods, for HeadingNet and the four baseline methods.}
  \label{fig:ae_plots_on_eval_baseline_and_NN}
\end{figure*}

\begin{figure}[!h]
    \centering
    \includegraphics[width=0.98\columnwidth]{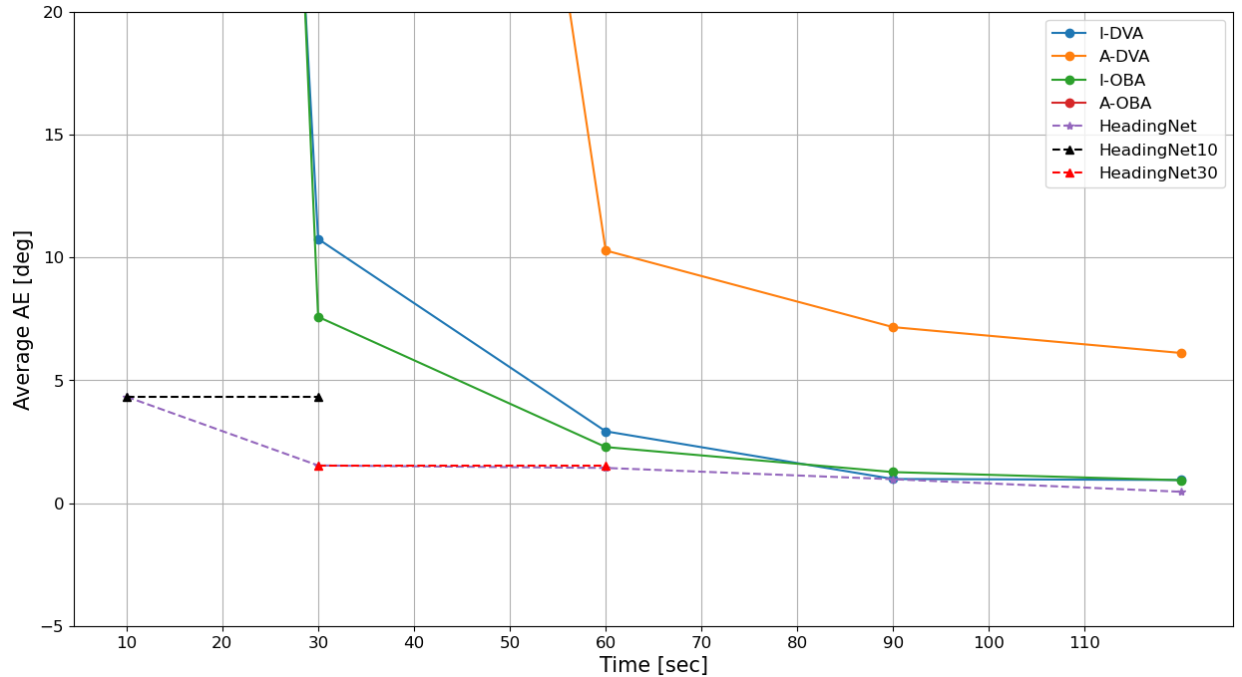}
    \caption{Average AE heading in degrees results of all baseline methods and HeadingNet variations. The black dashed line represents the heading AE of HeadingNet10, and the red dashed line represents the HeadingNet30 results, thus the proposed approach HeadingNet result at these two alignment times.}
    \label{fig:avg_ae_plots}
\end{figure}

\begin{figure}
    \centering
    \includegraphics[width=0.95\linewidth]{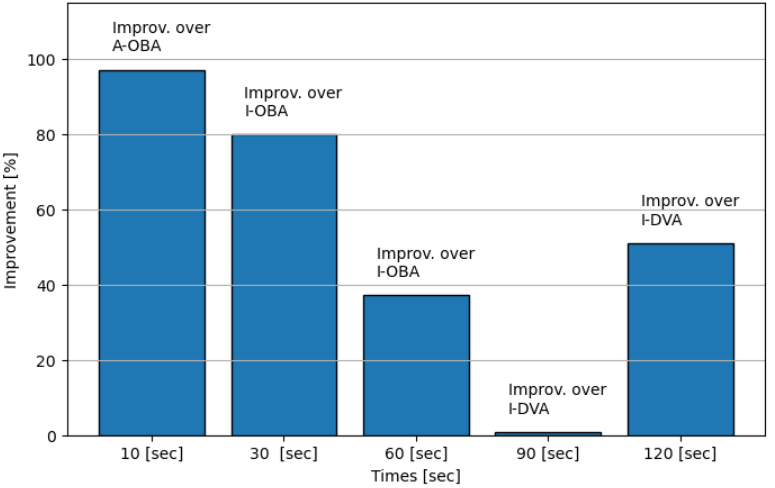}
    \caption{AE improvement bar plot in percentage showing our proposed the improvement of our proposed HeadingNet over the best baseline method.}
    \label{fig:improv_best_baseline}
\end{figure}

First, we analyze the AE of the heading angle estimation for the alignment times evaluated, for each of the four evaluation recordings. Figure \ref{fig:ae_plots_on_eval_baseline_and_NN} shows the AE of the heading angle for all four evaluation trajectories, EvalR1 to EvalR4.
Note that not all baseline approaches are presented in Figure \ref{fig:ae_plots_on_eval_baseline_and_NN}. This is because some baseline approaches achieved very poor AE results and remained out of scope even at the longest alignment time of 120 seconds. Therefore, they were omitted to improve the clarity of the results visualization. Examining the results in Figure \ref{fig:ae_plots_on_eval_baseline_and_NN}, it is evident that HeadingNet achieves significantly better performance for shorter alignment times. For the shortest alignment time of 10 seconds, HeadingNet demonstrates a significant improvement over all baseline approaches. Moreover, for evaluation recordings EvalR1, EvalR3, and EvalR4, the AE results of HeadingNet at the 10 second alignment time outperform the AE results of the baseline approaches at the longer 30 second alignment time. This provides an additional benefit beyond improved accuracy, namely faster convergence.\\
To fully evaluate the accuracy and time efficiency of HeadingNet against the baseline approaches, the AE results were averaged over the four evaluation recordings, EvalR1 to EvalR4, for each alignment time and for each method. Table \ref{average_ae_res_tbl} presents the averaged AE results for HeadingNet and the baseline approaches.
\begin{table}[h!]
\begin{adjustbox}{width = \columnwidth}
\begin{tabular}{|c|c|c|c|c|c|}
\hline
           & 10 Seconds $[\degree]$ & 30 Seconds $[\degree]$ & 60 Seconds $[\degree]$ & 90 Seconds $[\degree]$ & 120 Seconds $[\degree]$ \\ \hline
I-DVA      & 152.24     & 10.75      & 2.92       & 0.98       & 0.94        \\ \hline
A-DVA      & 154.13     & 88.19      & 10.28      & 7.16       & 6.11        \\ \hline
I-OBA      & 191.21     & 7.58       & 2.28       & 1.26       & 0.93        \\ \hline
A-OBA      & 150.49     & 114.6      & 79.23      & 77.83      & 38.31       \\ \hline
HeadingNet & \textbf{4.33 }      &\textbf{ 1.53 }      & \textbf{1.43}       & \textbf{0.97 }      & \textbf{0.46}        \\ \hline
\end{tabular}
\end{adjustbox}\caption{Average AE results of the heading angle in degrees. The average is performed for each alignment time over the four evaluation recordings, EvalR1 to EvalR4.}\label{average_ae_res_tbl}
\end{table}
From the table it can be observed that the AE of HeadingNet decreases with increasing alignment time, indicating convergence behavior as more data become available. In addition, HeadingNet outperforms all baseline methods for every evaluated alignment time, even for the longest alignment time of 120 seconds, where the model-based approaches are known to achieve their lowest AE values.\\
To further evaluate the accuracy relationship of HeadingNet compared to the baseline methods, Figure \ref{fig:avg_ae_plots} shows the average AE for each method and for each alignment time, based on the averaged AE results presented in Table \ref{average_ae_res_tbl}.\\
In the figure, two dashed lines are shown. The black dashed line indicates the HeadingNet AE at 10 seconds alignment time and is extended up to 30 seconds alignment time. The red dashed line shows the average HeadingNet AE at 30 seconds alignment time and is extended up to 60 seconds alignment time. These dashed lines illustrate that HeadingNet achieves lower AE at shorter alignment times than the baseline approaches achieve at longer alignment times, demonstrating faster convergence and indicating that the required alignment time can be reduced. For example, the AE of HeadingNet at 10 seconds alignment time is lower than that of all baseline approaches at 30 seconds alignment time, corresponding to a $66\%$ reduction in alignment time. Similarly, the AE of HeadingNet at 30 seconds alignment time is lower than that of all model-based approaches at 60 seconds alignment time, resulting in a $50\%$ reduction in alignment time. To further evaluate the AE accuracy and time efficiency of HeadingNet, Table \ref{improv_over_best_baseline_tbl} presents the average AE results of HeadingNet compared with the best performing model-based baseline approach for each alignment time, thereby providing a fair comparison with the established model-based approaches.
\begin{table}[h!]
\begin{adjustbox}{width = \columnwidth}
\begin{tabular}{|c|c|c|c|c|c|}
\hline
  & 10 secnods & 30 seconds & 60 seconds & 90 seconds & 120 seconds \\ \hline
Best baseline                                                      & A-OBA      & I-OBA      & I-OBA      & I-DVA      & I-DVA       \\ \hline
\begin{tabular}[c]{@{}c@{}}Best baseline\\ average AE\end{tabular} & 150.49     & 7.58       & 2.28       & 0.98       & 0.93        \\ \hline
\begin{tabular}[c]{@{}c@{}}HeadingNet\\ average AE\end{tabular}    & 4.33       & 1.53       & 1.43       & 0.97       & 0.46        \\ \hline
Improv.{[}\%{]}                                                    & 97.12      & 79.86      & 37.16      & 0.87       & 51.08       \\ \hline
\end{tabular}
\end{adjustbox}\caption{Comparison and improvement of our proposed approach, HeadingNet, with respect to the best performing baseline method.}\label{improv_over_best_baseline_tbl}
\end{table}\\
Two main conclusions can be drawn from Table \ref{improv_over_best_baseline_tbl}. First, HeadingNet consistently improves over the best baseline approach and therefore outperforms all baseline approaches. Second, the most significant improvements occur at shorter alignment windows ,10 and 30 seconds, while for the longest alignment time the improvement is smaller but still significant, reaching $51\%$ improvement in AE. To illustrate the AE improvement in percentage, Figure \ref{fig:improv_best_baseline} visualizes the relative improvement.\\
It is evident that HeadingNet achieves an average improvement of $53\%$ over the five evaluated alignment times. However, this average improvement is not reached at the 90 second alignment time. Nevertheless, at the longest alignment time of 120 seconds, HeadingNet still outperforms the best baseline approach, improving its AE by $51\%$. Moreover, if a very short alignment time is required, HeadingNet achieves a sub 5 degrees AE at the shortest alignment time of 10 seconds, thereby improving over the best baseline approach by $97\%$.
\section{Conclusion}\label{conc_sec}
Accurately estimating the heading angle of any autonomous platform such as an ASV is critical to initializing a precise navigation solution during its missions, consequently assuring the mission's success and validity. In this work, we address the prolonged alignment time required for model-based approaches, such as the vector and attitude decomposition methods, to achieve sufficient accurate alignment by proposing a neural-assisted framework, HeadingNet. HeadingNet is a data-driven multi-head neural network architecture designed for rapid and accurate heading estimation using only inertial measurements as input. HeadingNet's novelty is driven by a multi-head architecture, utilizing several two-dimensional convolutional neural network heads that processes inertial measurement to solve the attitude and provide heading estimation. Our proposed approach provides a simple end-to-end, model-free, in-motion self-heading alignment framework, that is able to significantly shorten the alignment time while increasing the accuracy. However, unlike the model-based approaches, which can provide alignment estimation at any selected time intervals, HeadingNet requires some modifications to the backbone architecture when applying to different alignment times.\\
HeadingNet was validated against state-of-the-art model-based approaches using real-world data recorded by an ASV in different mooring conditions. We show that HeadingNet is able to achieve an average improvement of $53\%$ compared to the best model-based approach, and shorten the alignment time by up to $67\%$. In addition, HeadingNet can achieve a sub-five degrees alignment error in just a ten second alignment time.\\
Overall, HeadingNet offers a model-free, end-to-end alignment framework which was validated using real-world data and showed significant improvement over the model-based approaches, therefore demonstrating its robustness, applicability, and overall benefits in terms of time and accuracy. HeadingNet is suitable for applications requiring rapid and accurate initial alignment, such as search and rescue operations or extended missions where on-the-fly alignment is essential.
\section*{Acknowledgment}
\noindent Z.Y. is supported by the Maurice Hatter Foundation and
University of Haifa presidential scholarship for outstanding
students on a direct Ph.D. track.



\bibliographystyle{IEEEtran}
\bibliography{bio.bib}

\end{document}